\documentclass{article}
\usepackage{spconf,amsmath,graphicx}

\usepackage{hyperref}

\usepackage{enumitem}
\setlist{nosep, leftmargin=*}
\usepackage[table]{xcolor}
\usepackage{amssymb, amsfonts}
\usepackage{euscript}
\usepackage{xcolor}
\usepackage{hyperref}
\usepackage{cleveref}
\usepackage{mathtools}
\usepackage{booktabs}
\usepackage{multirow}
\usepackage{threeparttable}
\usepackage{subcaption}
\usepackage{hyperref}
\graphicspath{Figures/}
\usepackage{graphicx}
\usepackage[export]{adjustbox}

\title{Explainable Transformer Prototypes for Medical Diagnoses\\
}

\name{Ugur Demir,  Debesh Jha,  Zheyuan Zhang,  Elif Keles,  Bradley Allen, Aggelos K. Katsaggelos, Ulas Bagci 
\thanks{This work is supported by NIH R01-CA246704, R01-CA240639, R03-EB032943, U01- DK127384-02S1, and U01-CA268808.}}
\address{Machine \& Hybrid Intelligence Lab, Northwestern University, Chicago, IL, USA}

\begin{document}
%
\maketitle

\begin{abstract}

Deployments of artificial intelligence in medical diagnostics mandate not just accuracy and efficacy but also \textit{trust}, emphasizing the need for explainability in machine decisions. The recent trend in automated medical image diagnostics leans towards the deployment of Transformer-based architectures, credited to their impressive capabilities. Since the self-attention feature of transformers contributes towards identifying crucial regions during the classification process, they enhance the trustability of the methods. However, the complex intricacies of these attention mechanisms may fall short of effectively pinpointing the regions of interest directly influencing AI decisions. Our research endeavors to innovate a unique attention block that underscores the correlation between 'regions' rather than 'pixels'. To address this challenge, we introduce an innovative system grounded in prototype learning, featuring an advanced self-attention mechanism that goes beyond conventional ad-hoc visual explanation techniques by offering comprehensible visual insights. A combined quantitative and qualitative methodological approach was used to demonstrate the effectiveness of the proposed method on the large-scale NIH chest X-ray dataset. Experimental results showed that our proposed method offers a promising direction for explainability, which can lead to the development of more trustable systems, which can facilitate easier and rapid adoption of such technology into routine clinics. Code is available at \url{www.github.com/NUBagcilab/r2r\_proto}. 
\end{abstract}

\begin{keywords}
Prototypes, Transformers,  Explainability, Interpretability, Medical Diagnosis
\end{keywords}

\section{Introduction}
\label{sec:intro}

\textbf{Explainability/Interpretability.} The interpretability of deep learning (DL) systems is of continuous concern within automated medical image diagnosis because the system decision can have negative consequences if the decision-making process is not clearly understood. This lack of transparency prevents a broader adoption and acceptance of these advanced technologies, calling for further work in the field of explainable AI. Thus, understanding this process is vital to developing trustable AI tools. Interpretable models provide insight into the internal mechanics while solving the problem. These details can include the steps taken to reach a conclusion or visualize the significant parts of the input data. Visualizing the input region that the model uses for decision-making can be beneficial for reducing the workload of the operators. 

\textbf{Visual explanations.} Recent developments in the field of interpretable DL have led to two fundamental approaches. The first group of methods obtains ad-hoc explanations from the existing deep neural networks (DNN) by using intermediate activation maps and gradient information such as CAM~\cite{cam} and famously Grad-CAM~\cite{gradcam} \cite{HarmonArtificial2020}. Ad-hoc explanations generally do not change the model architecture or the predictions. Despite their widespread use, they tend to be unstable and sensitive to small changes in the input space~\cite{NEURIPS2019_fooling_interpretations}. This problem is partially mitigated by integrating the \textit{information bottleneck} in the visualization step~\cite{iba}~\cite{demir_iba}; it makes the process relatively slower, however raising efficiency concerns. 

\textbf{Model interpretation.} Another popular approach for model interpretation is designing the DNN architecture to be more transparent~\cite{Bohle_2021_CVPR}~\cite{Nauta_2021_CVPR}. Arguably, altering the model architecture for the benefit of interpretation may lead to performance drops~\cite{Lim_2021_CVPR_ReliableExplanations} but it is largely based upon empirical studies with opposite results started to appear too such that a more carefully designed interpretation modules can fill the performance gap~\cite{xprotonet}.


\textbf{Prototype learning.} Using interpretable layers with the existing CNN architectures can be beneficial thanks to the availability of the pre-trained weights. In~\cite{xprotonet}, for instance, the model uses ResNet~\cite{resnet} architecture as a backbone for feature extraction~\cite{convproto}. An additional branch uses these features to predict masks. The global averaging operation uses mask weights to highlight important spatial regions and removes unnecessary information. The similarity between the masked features and the parameterized prototype vectors to perform classification. The mask associated with the most similar prototype is used as a visual explanation. Although the results show promising performance on NIH Chest X-Ray~\cite{nih_xray} dataset, the proposed interpretation module can only be applied once after extracting the convolutional features. It limits the resolution of the obtained masks since each layer in the backbone architecture has a certain spatial resolution.

\textbf{Transformers.} Transformer architecture has been successfully used for vision tasks for different problems~\cite{dosovitskiy2020vit}~\cite{chen2021transunet}. At the heart of its efficacy lies the multi-head self-attention mechanism, which excels at discerning long-range relationships within the input space. Attention maps derived from this mechanism can help spotlight the most salient regions. However, the self-attention mask, in its present form, offers no assurance of effective information masking. This limits its utility in providing trustworthy and comprehensible explanations for AI decisions. A potential solution for this quandary is to use of an interpretable self-attention module. Such an enhancement could help unpacking the 'black box' nature of Transformer models, bringing a new level of transparency and understanding to these powerful tools. 

\textbf{What do we propose?} In our study, we present a new prototype learning algorithm which seamlessly integrates a self-attention layer into the Transformer architecture. Our method leverages the Convolutional Vision Transformer (CvT) architecture to harness the power of convolutional features and maintain smaller parameter sizes. Our innovative self-attention layer uses additional masking branches and a global average pooling strategy to derive\textit{ 'query'} vectors. A key aspect of this formulation is that it doesn't condition \textit{keys} and \textit{values} on the input. Instead, key and value vectors are stored as 'parametric prototypes' and remain unconditioned on the input features.  This greatly allows us to use the new region-to-region attention instead of pixel-to-pixel.  Each query vector exclusively contains information about the regions indicated by the corresponding mask, and the attention matrix illustrates the similarity between these regions. Then, the attention and the masks are integrated (multiplied) to highlight important masks. We utilized the mask to find out the most active mask index to select the corresponding value vector to form the output features. The proposed architecture replaces all of the self-attention layers in CvT. This allows us to obtain explanation masks from different resolution levels.


The main contributions our study can be summarized as;
\begin{itemize}
\item \textbf{Interpretable self-attention mechanism:} We propose a novel interpretable self-attention mechanism within a prototype learning paradigm with vision transformers. Our \textit{region-to-region self-attention} replaces the traditional grid-based patch-splitting operation with parametric prototype representation learning. It provides significantly better visual explanations. 


\item \textbf{Rigorous evaluation:} The effectiveness of our proposed architecture is first demonstrated on NIH Chest X-Ray dataset~\cite{wang2017chestx} as an example high-risk AI application. 

 \item \textbf{Explainablity and interpretability:} Our novel method has a self-attention mechanism that can produce visual explanation at each intermediate feature layer for different resolutions, improving explainability and diagnostic performance.

\end{itemize}

\begin{figure*}[t]
\centering
\includegraphics[width=1.9\columnwidth]{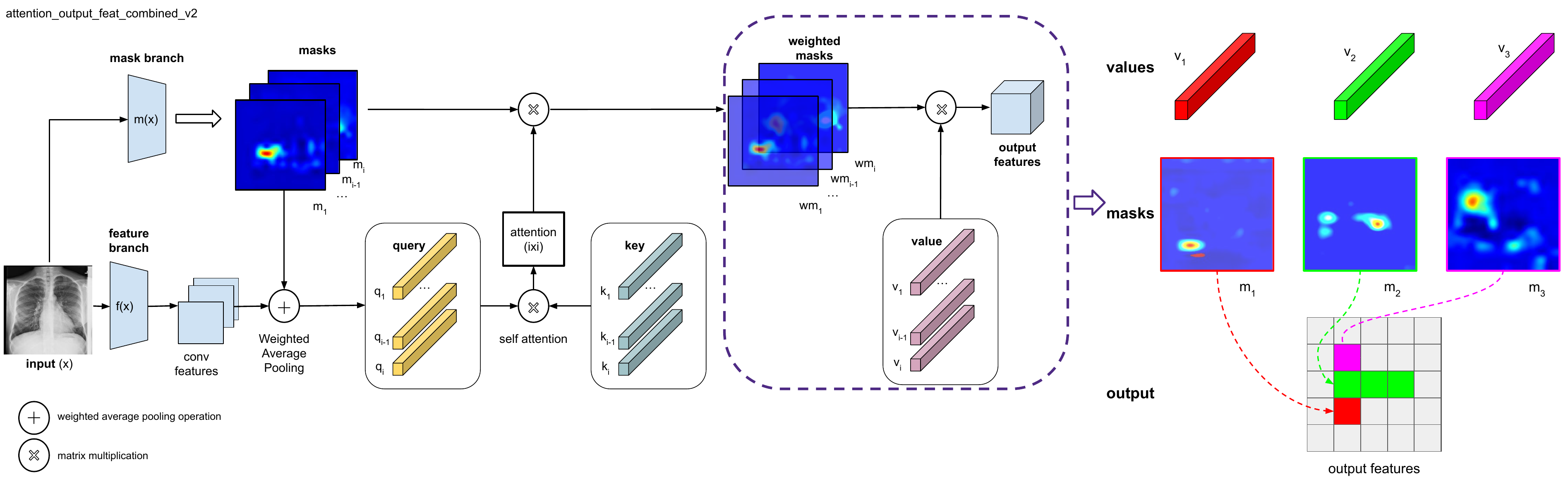} 
\caption{Interpretable self-attention architecture. It predicts masks to highlight important regions in the input. The feature branch extracts convolutional features from the input to form query vectors. The predicted masks are applied to the convolutional features to remove unnecessary information. The attention values are calculated by multiplying query and key vectors. Finally, value vectors and the weighted masks are multiplied to obtain output features.}
\label{fig:attn}
\end{figure*}

\section{Method}
Figure \ref{fig:attn} depicts the overall architecture of the proposed interpretable self-attention. This module replaces the attention layers in CvT architecture. It consists of three main parts: (i) mask and query generation, (ii) region-to-region self-attention, and (iii) output feature reconstruction. In the following section, each part of the proposed interpretable self-attention module is introduced.


\subsection{Part I: Mask and Query Generation}
The standard transformer network splits the input image into small patches and the attention module tries to find patch-level similarities in the input features. The patching operation uses standard grid lines to determine patch borders. The location information is preserved by attaching positional encoding to the patch features. This approach limits the effectiveness of attention since patches can contain parts from the background and foreground. We introduce a novel masking-based query generation to implement adaptive patch generation. The masking operation $M$


\begin{equation}
    m_i = \Theta(M_i(x))
\end{equation}
generates a certain number of masks (i.e., pre-defined) for the given input. The $\Theta$ performs Softmax operation to enforce only one mask is dominant in each pixel location. 

The feature extraction branch calculates the features $f$ by
\begin{equation}
    f = F(x)
\end{equation}
where $F$ is convolutional projection implemented by depth-wise separable convolution \cite{depthconv}. Features are then pooled with masks to transform convolutional features into queries. We multiply each mask $m_i$ with feature $f$ to restrict related information only into a specific mask. In other words, feature tensor $f$  does not include pixel information when mask values are zero. The masked convolutional features are summed in spatial dimension to obtain the query vector
\begin{equation}
    q_i^c = \sum_{h,w} m_i^{hw} \odot f^{hwc}
\end{equation}
where $h$, $w$, $c$ represents height, width, channel dimensions and $\odot$ is element-wise multiplication. This operation can be considered as weighted global average pooling where the importance weights are determined by the mask. The query vector $q_i$ represents the image region located by the mask. This gives flexible patch selection compared to standard gridline splitting. Since each $q_i$ is associated with the mask $m_i$, we can consider the mask as position embedding. 



\begin{figure*}
\centering
\includegraphics[width=2.0\columnwidth]{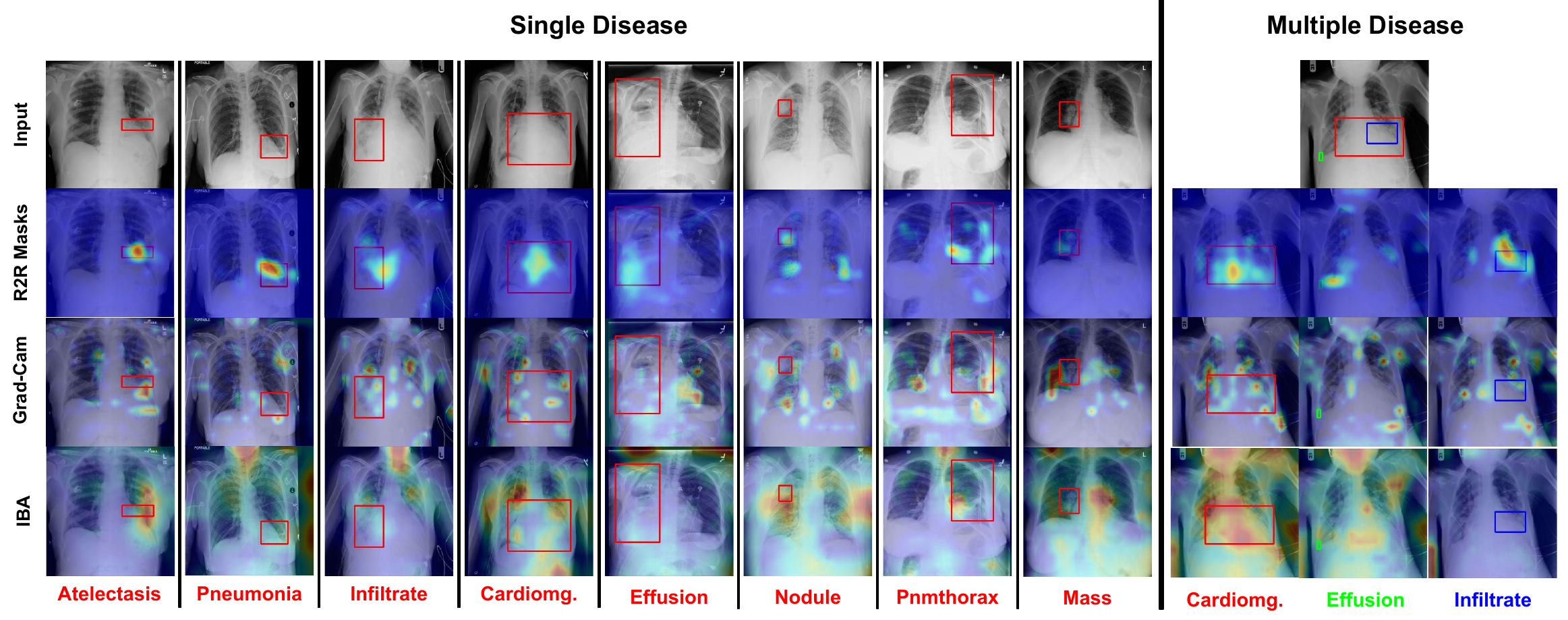}
\caption{Example mask visualizations of the proposed method on the NIH Chest X-Ray~\cite{wang2017chestx} dataset. We selected the most active masks for visualization.}
\label{fig:masks_xray}
\end{figure*}



\begin{table*}
\small
\centering
\caption{Results on NIH Chest X-Ray dataset. The models with * are trained from scratch in our framework. The column names represent abbreviations for disease names in the order of Atelectasis, Cardiomegaly, Effusion, Infiltration, Mass, Nodule, Pneumonia, Pneumothorax, Consolidation, Edema, Emphysema, Fibrosis, Pleural Thickening, Hernia.}
\tabcolsep=0.08cm
\begin{tabular}{|l|c|c|c|c|c|c|c|c|c|c|c|c|c|c||c|}
\hline
\textbf{Method} & \textbf{Atel} & \textbf{Card} & \textbf{Effu} & \textbf{Infi} & \textbf{Mass} & \textbf{Nodu} & \textbf{Pne1} & \textbf{Pne2} & \textbf{Cons} & \textbf{Edem} & \textbf{Emph} & \textbf{Fibr} & \textbf{P.T.} & \textbf{Hern}& \textbf{Mean} \\
\hline
Wang et al.\cite{nih_xray} & 0.700 & 0.810 & 0.759 & 0.661 & 0.693 & 0.669 & 0.658 & 0.799 & 0.703 & 0.805 & 0.833 & 0.786 & 0.684 & 0.872 & 0.745 \\
Guan et al. (ResNet)\cite{GUAN2020259}  & 0.779 & 0.879 & 0.824 & 0.694 & 0.831 & 0.766 & 0.726 & 0.858 & 0.758 & 0.850 & 0.909 & 0.832 & 0.778 & 0.906 & 0.814 \\
Guan et al. (DenseNet)\cite{GUAN2020259} & 0.781 & 0.883 & 0.831 & 0.697 & 0.830 & 0.764 & 0.725 & 0.866 & 0.758 & 0.853 & 0.911 & 0.826 & 0.780 & 0.918 & 0.816 \\
Ma et al.\cite{Ma2019Multi} & 0.777 & 0.894 & 0.829 & 0.696 & 0.838 & 0.771 & 0.722 & 0.862 & 0.750 & 0.846 & 0.908 & 0.827 & 0.779 & 0.934 & 0.817 \\
Hermoza et al.\cite{hermoza2020region}         & 0.775 & 0.881 & 0.831 & 0.695 & 0.826 & 0.789 & 0.741 & 0.879 & 0.747 & 0.846 & 0.936 & 0.833 & 0.793 & 0.917 & 0.821 \\
Kim et al. (ResNet)\cite{xprotonet} & 0.782 & 0.881 & 0.836 & 0.715 & 0.834 & 0.799 & 0.730 & 0.874 & 0.747 & 0.834 & 0.936 & 0.815 & 0.798 & 0.896 & 0.820 \\
Kim et al. (DenseNet)\cite{xprotonet}   & 0.780 & 0.887 & 0.835 & 0.710 & 0.831 & 0.804 & 0.734 & 0.871 & 0.747 & 0.840 & 0.941 & 0.815 & 0.799 & 0.909 & 0.822 \\
\hline   \rowcolor[rgb]{1,0.95,0.88}
CvT* \cite{cvt} & 0.663 & 0.808 & 0.740 & 0.614 & 0.660 & 0.591 & 0.590 & 0.688 & 0.624 & 0.755 & 0.697 & 0.686 & 0.665 & 0.773 & 0.682 \\
 \rowcolor[rgb]{.9,0.95,1}
 Ours & 0.752 & 0.873 & 0.817 & 0.696 & 0.784 & 0.738 & 0.699 & 0.850 & 0.734 & 0.837 & 0.893 & 0.807 & 0.765 & 0.928 & 0.798\\
\hline
\end{tabular}

\label{table:xray}
\end{table*}

\subsection{Part II: Region-to-Region Self-Attention}
In order to find self-similarities and calculate the attention values, we multiply queries ($q$) with keys ($k$). In the standard self-attention, the keys are calculated by projecting input features using linear or convolution layers. In our module, we adopted an alternative approach to defining keys by using parametric learnable prototype vectors. As depicted in Figure \ref{fig:attn}, key vectors are not conditioned on the input; instead, the model stores learnable key vectors $k_i$. This formulation leads the model to learn canonical representation for important patterns. For each mask $m_i$ and query $q_i$, we define a key vector $k_i$. The self-attention is calculated by matrix multiplication
\begin{equation}
    attn_{L \times L} = q_{L \times d} \otimes k_{L \times d}^T
\end{equation}
where $L$ is the number of masks, $d$ is the dimension of query and key vectors, $q_{L \times d}$ and $k_{L \times d}$ are $L \times d$ matrices. The attention matrix $attn$ has $L \times L$ dimension and indicates the correlation between the query and key vectors. We use the attention matrix as weighting coefficients to emphasize more dominant masks. Figure \ref{fig:attn} depicts the weighed masks $wm_i$ by changing their opacity. 

This aspect of the attention approach can be viewed as \textit{region-to-region self-attention}. We obtain local regions dynamically while calculating query vectors. The main objective of the masking branch is to find a segment of the input that has a similar pattern to the corresponding key vector $k_i$. Keys act as prototype vectors and learn common patterns in the datasets.


\subsection{Part III: Output Feature Reconstruction}

The last step of the attention module focuses on reconstructing the output features for the next layer. We define parametric learnable prototype value vectors $v_i$. At this stage, weighted masks $wm_i$ have the location information about the canonical object part locations. We can form a feature matrix by placing similar value vectors $v_i$ to pixels if they are in the same mask $wm_i$. This can be achieved by another matrix multiplication
\begin{equation}
    o_{z \times hw} = v_{z \times L} \otimes wm_{L \times hw}
\end{equation}
where $z$ is the dimension of each value vector $v_i$, $L$ is the number of masks with spatial dimension $h \times w$ and $o$ is the output feature map. 

\section{Experiments and Results}
\textbf{Datasets.} We used public NIH chest X-ray dataset~\cite{wang2017chestx}. It consists of 112,120 frontal-view X-ray images having 14 different types of disease labels obtained from 30,805 unique patients. 

\textbf{Implementation details.} We used the PyTorch library to implement our training framework. For all of the experiments, the AdamW optimizer is used with $0.00025$ learning rate and $0.05$ weight decay. We utilized the Cosine Annealing learning rate scheduler for learning rate updates. For the multi-label classifications, binary cross entropy is used. We used CvT-13 configuration for the baseline experiments. 


\textbf{Quantitative results.} In the initial step, we conducted our experiments on natural image datasets. The aim of this study is to test our new interpretable self-attention design against the baseline CvT architecture. It is important to note that we trained our models from scratch without using any pre-trained weights. We used the same hyper-parameters and learning strategy for all of the methods. At this stage of the study, we avoided extensive hyper-parameter searches. 

Table \ref{table:xray} shows the comparison between our method and the state-of-the-art methods on the same benchmark. We used the AUC score as a comparison metric. The table shows AUC scores for each disease separately and the mean AUC score in the last column. The XprotoNet \cite{xprotonet} has the best performance of 0.822 AUC score when it uses DenseNet with the pre-trained ImageNet-1000 weights as a feature extractor. We achieved 0.798 AUC score without using any pre-training. Our from-scratch training is able to perform on par with the SOTA methods. This performance gap can be closed by pre-training on large datasets.

\textbf{Qualitative results.} This section demonstrates the mask visualizations from the interpretable self-attention module. Each stage self-attention module has multiple mask predictions. For each mask generation stage, we only have a few active mask predictions. This makes it easy to select useful masks from the large set of generated masks. We only demonstrated the most active masks in this section. We compared our heatmaps against the renowned Grad-CAM explanations and the Information Bottleneck Attribution (IBA) techniques. The visual outcomes for patients diagnosed with singular and multiple diseases are presented in Figure \ref{fig:masks_xray}, pinpointing the lesion's location corresponding to the specified disease. Figure \ref{fig:masks_xray} illustrates cases with patients having multiple diseases. The provided heatmaps demonstrate the impact of the weight of the network. The red and yellow regions of the heatmaps denote the \textit{most relevant features} whereas the blue regions show the \textit{least relevant features}. The heatmaps show the interpretability of the proposed method. In both scenarios, our region-to-region attention more accurately identifies the lesion locations.




\section{Conclusion}
A standout feature of our work is the novel interpretable self-attention mechanism, which offers visual explanations for disease, a leap forward from previous methodologies. Both quantitative and qualitative results provide strong support for our method's visual explainability, outperforming or performing on-par ad-hoc explanation methods of the past. Our proposed method even surpassed the performance of CvT-13, a recent and robust benchmark algorithm in medical image classification tasks. 
In the future, we will explore the robustness of the proposed algorithm both in medical and non-medical domains. 




\bibliographystyle{IEEEtran}
\bibliography{IEEEabrv,isbi2024}

\begin{thebibliography}{10}
\providecommand{\url}[1]{#1}
\csname url@samestyle\endcsname
\providecommand{\newblock}{\relax}
\providecommand{\bibinfo}[2]{#2}
\providecommand{\BIBentrySTDinterwordspacing}{\spaceskip=0pt\relax}
\providecommand{\BIBentryALTinterwordstretchfactor}{4}
\providecommand{\BIBentryALTinterwordspacing}{\spaceskip=\fontdimen2\font plus
\BIBentryALTinterwordstretchfactor\fontdimen3\font minus
  \fontdimen4\font\relax}
\providecommand{\BIBforeignlanguage}[2]{{%
\expandafter\ifx\csname l@#1\endcsname\relax
\typeout{** WARNING: IEEEtran.bst: No hyphenation pattern has been}%
\typeout{** loaded for the language `#1'. Using the pattern for}%
\typeout{** the default language instead.}%
\else
\language=\csname l@#1\endcsname
\fi
#2}}
\providecommand{\BIBdecl}{\relax}
\BIBdecl

\bibitem{cam}
B.~Zhou, A.~Khosla, L.~A., A.~Oliva, and A.~Torralba, ``{Learning Deep Features
  for Discriminative Localization.}'' \emph{CVPR}, 2016.

\bibitem{gradcam}
R.~R. Selvaraju, M.~Cogswell, A.~Das, R.~Vedantam, D.~Parikh, and D.~Batra,
  ``Grad-cam: Visual explanations from deep networks via gradient-based
  localization,'' in \emph{Proceedings of the IEEE International Conference on
  Computer Vision (ICCV)}, 2017.

\bibitem{HarmonArtificial2020}
S.~A. Harmon, T.~H. Sanford, S.~Xu, E.~B. Turkbey, H.~Roth, Z.~Xu, D.~Yang,
  A.~Myronenko, V.~Anderson, A.~Amalou \emph{et~al.}, ``Artificial intelligence
  for the detection of covid-19 pneumonia on chest ct using multinational
  datasets,'' \emph{Nature communications}, vol.~11, no.~1, pp. 1--7, 2020.

\bibitem{NEURIPS2019_fooling_interpretations}
J.~Heo, S.~Joo, and T.~Moon, ``Fooling neural network interpretations via
  adversarial model manipulation,'' in \emph{Advances in Neural Information
  Processing Systems}, H.~Wallach, H.~Larochelle, A.~Beygelzimer,
  F.~d\textquotesingle Alch\'{e}-Buc, E.~Fox, and R.~Garnett, Eds., vol.~32,
  2019.

\bibitem{iba}
K.~Schulz, L.~Sixt, F.~Tombari, and T.~Landgraf, ``Restricting the flow:
  Information bottlenecks for attribution,'' in \emph{Proceedings of the
  International Conference on Learning Representations}, 2020.

\bibitem{demir_iba}
U.~Demir, I.~Irmakci, E.~Keles, A.~Topcu, Z.~Xu, C.~Spampinato,
  S.~Jambawalikar, E.~Turkbey, B.~Turkbey, and U.~Bagci, ``Information
  bottleneck attribution for visual explanations of diagnosis and prognosis,''
  in \emph{MLMI 2021}, vol. 12966, 2021, pp. 396--405.

\bibitem{Bohle_2021_CVPR}
M.~Bohle, M.~Fritz, and B.~Schiele, ``Convolutional dynamic alignment networks
  for interpretable classifications,'' in \emph{Proceedings of the IEEE/CVF
  Conference on Computer Vision and Pattern Recognition (CVPR)}, 2021, pp.
  10\,029--10\,038.

\bibitem{Nauta_2021_CVPR}
M.~Nauta, R.~van Bree, and C.~Seifert, ``Neural prototype trees for
  interpretable fine-grained image recognition,'' in \emph{Proceedings of the
  IEEE/CVF Conference on Computer Vision and Pattern Recognition (CVPR)}, 2021,
  pp. 14\,933--14\,943.

\bibitem{Lim_2021_CVPR_ReliableExplanations}
D.~Lim, H.~Lee, and S.~Kim, ``Building reliable explanations of unreliable
  neural networks: Locally smoothing perspective of model interpretation,'' in
  \emph{CVPR}, 2021, pp. 6468--6477.

\bibitem{xprotonet}
E.~Kim, S.~Kim, M.~Seo, and S.~Yoon, ``Xprotonet: Diagnosis in chest
  radiography with global and local explanations,'' in \emph{Proceedings of the
  IEEE/CVF Conference on Computer Vision and Pattern Recognition (CVPR)}, 2021,
  pp. 15\,719--15\,728.

\bibitem{resnet}
A.~{Chattopadhay}, A.~{Sarkar}, P.~{Howlader}, and V.~N. {Balasubramanian},
  ``Grad-cam++: Generalized gradient-based visual explanations for deep
  convolutional networks,'' in \emph{Proceedings of the IEEE Winter Conference
  on Applications of Computer Vision (WACV)}, 2018, pp. 839--847.

\bibitem{convproto}
H.-M. Yang, X.-Y. Zhang, F.~Yin, and C.-L. Liu, ``Robust classification with
  convolutional prototype learning,'' in \emph{CVPR}, June 2018.

\bibitem{nih_xray}
X.~Wang, Y.~Peng, L.~Lu, Z.~Lu, M.~Bagheri, and R.~M. Summers, ``Chestx-ray8:
  Hospital-scale chest x-ray database and benchmarks on weakly-supervised
  classification and localization of common thorax diseases,'' in \emph{CVPR},
  July 2017.

\bibitem{dosovitskiy2020vit}
A.~Dosovitskiy, L.~Beyer, A.~Kolesnikov, D.~Weissenborn, X.~Zhai,
  T.~Unterthiner, M.~Dehghani, M.~Minderer, G.~Heigold, S.~Gelly, J.~Uszkoreit,
  and N.~Houlsby, ``An image is worth 16x16 words: Transformers for image
  recognition at scale,'' \emph{ICLR}, 2021.

\bibitem{chen2021transunet}
J.~Chen, Y.~Lu, Q.~Yu, X.~Luo, E.~Adeli, Y.~Wang, L.~Lu, A.~L. Yuille, and
  Y.~Zhou, ``Transunet: Transformers make strong encoders for medical image
  segmentation,'' \emph{arXiv preprint arXiv:2102.04306}, 2021.

\bibitem{wang2017chestx}
X.~Wang, Y.~Peng, L.~Lu, Z.~Lu, M.~Bagheri, and R.~M. Summers, ``Chestx-ray8:
  Hospital-scale chest x-ray database and benchmarks on weakly-supervised
  classification and localization of common thorax diseases,'' in \emph{CVPR},
  2017, pp. 2097--2106.

\bibitem{depthconv}
F.~Chollet, ``Xception: Deep learning with depthwise separable convolutions,''
  in \emph{2017 IEEE Conference on Computer Vision and Pattern Recognition
  (CVPR)}, 2017, pp. 1800--1807.

\bibitem{GUAN2020259}
Q.~Guan and Y.~Huang, ``Multi-label chest x-ray image classification via
  category-wise residual attention learning,'' \emph{Pattern Recognition
  Letters}, vol. 130, pp. 259--266, 2020.

\bibitem{Ma2019Multi}
C.~Ma, H.~Wang, and S.~Hoi, \emph{Multi-label Thoracic Disease Image
  Classification with Cross-Attention Networks}, 2019, pp. 730--738.

\bibitem{hermoza2020region}
R.~Hermoza, G.~Maicas, J.~C. Nascimento, and G.~Carneiro, ``Region proposals
  for saliency map refinement for weakly-supervised disease localisation and
  classification,'' in \emph{MICCAI}, 2020, pp. 539--549.

\bibitem{cvt}
H.~Wu, B.~Xiao, N.~Codella, M.~Liu, X.~Dai, L.~Yuan, and L.~Zhang, ``Cvt:
  Introducing convolutions to vision transformers,'' in \emph{ICCV}, October
  2021, pp. 22--31.

\end{thebibliography}

\end{document}